\documentclass[conference]{IEEEtran}
\IEEEoverridecommandlockouts
\usepackage{cite}
\usepackage{amsmath,amssymb,amsfonts}
\usepackage{algorithmic}
\usepackage{graphicx}
\usepackage{textcomp}
\usepackage{xcolor}

\usepackage{booktabs}
\usepackage{multirow}

\def\BibTeX{{\rm B\kern-.05em{\sc i\kern-.025em b}\kern-.08em
    T\kern-.1667em\lower.7ex\hbox{E}\kern-.125emX}}
\begin{document}

\title{Policy-Driven CT-Agent: Modeling Phase-Aware Diagnostic Control for Clinically Consistent CT Reasoning\\
% {\footnotesize \textsuperscript{*}Note: Sub-titles are not captured for https://ieeexplore.ieee.org  and
}

% \author{Anonymous}

\author{
{Yanmeng Dong} \textsuperscript{1},  
{Han Li} \textsuperscript{2},  
{Yujia Li} \textsuperscript{1},  
{Jingsong Liu} \textsuperscript{2},
{Xun Ma} \textsuperscript{2},  
{Yanzhu Hu} \textsuperscript{2},  
{Zhengyang Xu} \textsuperscript{2}, \\
{Zhicheng Li} \textsuperscript{1}, 
{Nassir Navab} \textsuperscript{2}, 
{Shaohua Kevin Zhou} \textsuperscript{1}   \thanks{Corresponding author.} \\
\textsuperscript{1}University of Science and Technology of China
\hspace{0.2cm}\textsuperscript{2}Technical University of Munich
}

\maketitle

\begin{abstract}
Computed Tomography (CT) diagnosis often relies on dynamic selection of imaging phases, such as non-contrast, arterial, or venous phases, based on preliminary findings, clinical suspicion, and diagnostic guidelines. This phase-wise decision process is critical for reducing unnecessary radiation exposure while supporting timely staging and treatment planning. However, phase-selection protocols can vary across hospitals, regions, and guidelines, while most existing CT-based AI methods assume that all phases are available and focus on static tasks under a fixed imaging phase, failing to \textbf{model whether additional phases are required.}
This limitation stems from heterogeneous multi-phase representations, the need for knowledge-guided phase control beyond visual cues, and the lack of supervision for phase-sufficiency decisions in existing datasets. To address these challenges, we propose \textbf{Policy-Driven CT-Agent (PD-CTAgent)} for clinically consistent CT phase selection and diagnostic reasoning. PD-CTAgent introduces a Clinical Structure Abstraction Module (CSAM) to harmonize heterogeneous CT phases into a unified, phase-aware evidence representation. Based on this representation, a Knowledge-Guided Diagnostic Control Model (KDCM) evaluates phase sufficiency and iteratively requests additional phases when necessary. The policy-driven agent design further allows PD-CTAgent to flexibly follow different institutional, regional, or guideline-specific diagnostic protocols.
Together, PD-CTAgent bridges static CT analysis and real-world clinical workflows. Experiments on two public datasets, LIDC and MCT-LTDiag, and one private dataset demonstrate its effectiveness and clinical consistency. Code will be made public upon acceptance.
\end{abstract}

\begin{IEEEkeywords}
Multi-Phase CT, Policy-Driven Reasoning, Medical Agents.
\end{IEEEkeywords}

\begin{figure*}[t]
    \centering
    \includegraphics[width=0.87\textwidth]{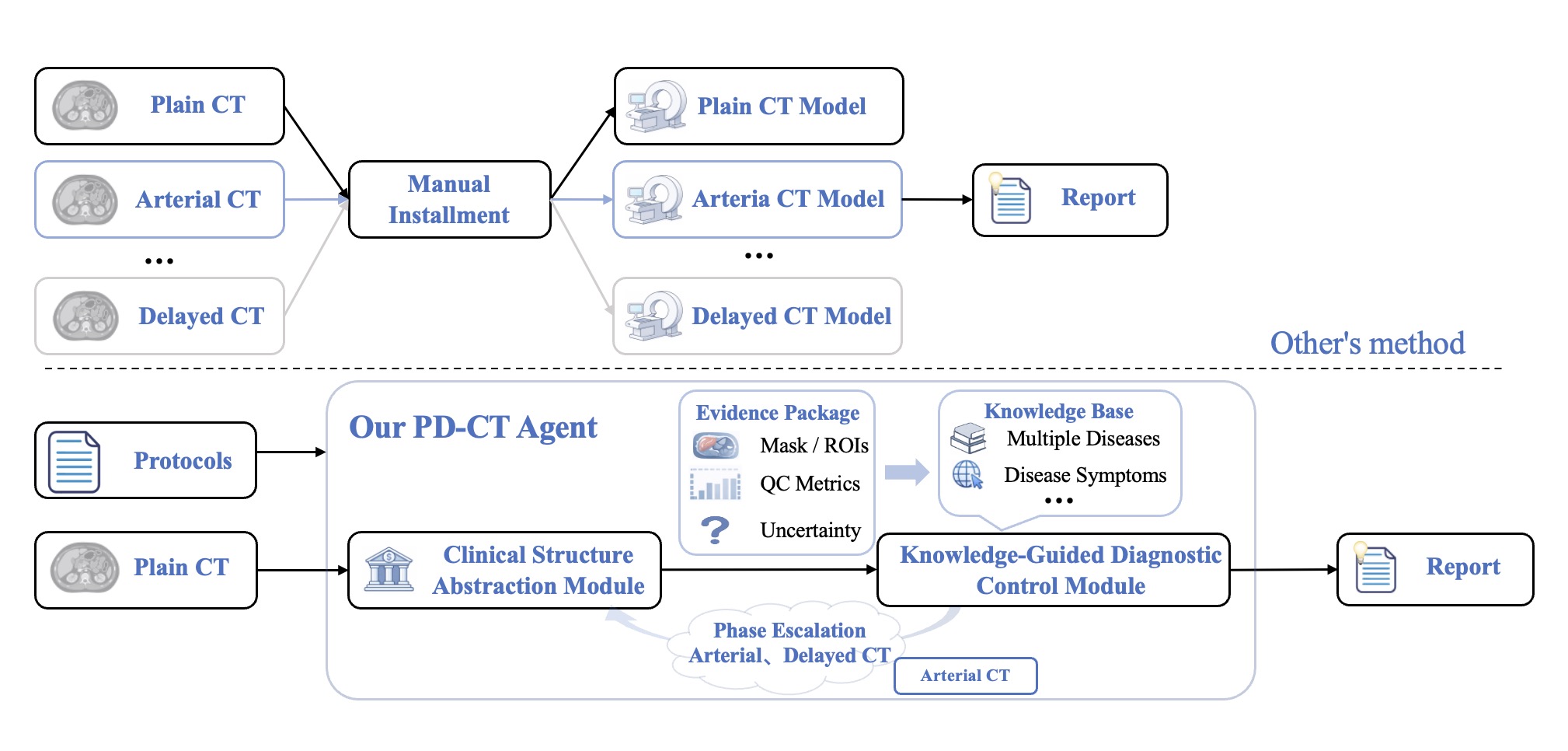}
    \vspace{-0.6cm}
    \caption{Comparison between conventional CT diagnostic pipelines and our PD-CT Agent. Traditional methods rely on manually selected existing phase-specific models (e.g., arterial and delayed CT), whereas our PD-CT Agent starts from plain CT, performs adaptively escalates phases to generate the final report.}
    \label{fig:teaser}
    \vspace{-0.3cm}
\end{figure*}

\section{Introduction}
Computed Tomography (CT) plays a central role in the diagnosis and staging of abdominal and thoracic diseases~\cite{zhong2026abn,chen2024mask}.
In real-world clinical workflows, radiologists dynamically select CT phases based on preliminary findings, clinical suspicion, and diagnostic protocols, which can involve whether to perform a plain or contrast-enhanced scan, as well as which contrast-enhanced phase, arterial, venous, or excretory, should be acquired~\cite{zhang2025dct,zhang2025multimodal}.
Importantly, such phase-selection strategies are not always uniform, as diagnostic guidelines and institutional protocols may vary across hospitals, regions, and clinical scenarios.
For example, kidney lesions are often initially detected on non-contrast scans, while enhancement patterns observed in arterial and venous phases are critical for determining malignancy~\cite{van2023incidental,chu2020protocol}.
This phase-wise decision process is clinically essential: benign findings may not require additional contrast-enhanced imaging, reducing costs and radiation exposure, whereas suspected malignancies require immediate staging and treatment planning rather than delayed decisions driven by additional imaging~\cite{bucolo2023virtual,o2011incidental}.

Despite its clinical importance, most existing CT-based AI methods focus on static tasks such as lesion detection, segmentation, or classification \textbf{under a fixed imaging phase}~\cite{ling2023mtanet,kong2024enhancing,ahmad2023revolutionizing,jiang2023deep}.
They typically assume that all CT phases are already available and do not explicitly model the decision process of whether additional phases are required, as illustrated in Fig.~\ref{fig:teaser}.
This limitation is not merely a design choice, but stems from inherent technical challenges:
\textbf{(1) multi-phase comprehension poses substantial difficulty.}
Different CT phases exhibit distinct contrast distributions, anatomical emphasis, and diagnostic roles.
Effectively integrating heterogeneous phase-specific evidence into a unified representation space requires cross-phase semantic alignment, which is non-trivial for conventional vision models.
\textbf{(2) knowledge-guided control is fundamentally challenging.}
Clinical decisions about requesting additional CT phases rely on structured diagnostic knowledge, institutional protocols, and experience-based policies rather than purely visual cues.
Moreover, such phase-selection policies may vary across hospitals, regions, and guideline-specific diagnostic settings.
Encoding such decision policies into computational models remains underexplored, and existing datasets do not explicitly model phase-wise diagnostic sufficiency or escalation decisions.
These challenges largely explain why current CT AI systems remain disconnected from real-world diagnostic workflows. Importantly, the clinical question is not to replace CT acquisition protocols or to determine scanning strategies autonomously. Instead, we focus on a complementary problem: given partially observed CT evidence, whether the current information is sufficient for diagnosis or additional evidence (e.g., extra phases or structured clinical cues) is required.

Recent advances in Vision-Language Models (VLMs) and Large Language Models (LLMs) provide new opportunities for structured diagnostic reasoning~\cite{mao2025ct,ben2024cpllm,leonardi2023enhancing,li2024agent,shi2024medadapter,ghezloo2025pathfinder,wang2025m3builder}.
LLMs demonstrate strong capabilities in knowledge integration and decision modeling, while VLMs enable joint reasoning over visual evidence and clinical context~\cite{liu2025medchat}. These developments enable moving beyond static image analysis toward workflow-aware, policy-driven diagnostic modeling.

However, these models are still inherently static: they take a fixed input and produce a single-step prediction, without explicitly modeling the sequential process of evidence evaluation, evidence acquisition, and iterative reassessment. This limitation motivates an agent-based formulation that maintains explicit diagnostic state and enables iterative interaction with evolving clinical evidence.

Motivated by this opportunity, as shown in Fig.~\ref{fig:teaser}, we propose \textbf{Policy-Driven CT-Agent (PD-CTAgent)}, an agent-based framework that models clinically consistent CT phase selection and diagnostic reasoning.
To address the above limitations, we design PD-CTAgent as an interactive diagnostic agent with explicit state tracking and iterative evidence refinement.
PD-CTAgent introduces two key components:
\textbf{(1) Clinical Structure Abstraction Module (CSAM).}
As described in Fig.~\ref{fig:pdct_agent} (a), to address multi-phase comprehension, we design a structured abstraction mechanism for heterogeneous CT phases.
For each CT phase, organ-level segmentation and uncertainty estimation are performed.
Phase-specific prompts describing diagnostic roles are attached, and all information is encapsulated into a structured \emph{Evidence Package}.
The Evidence Package is simultaneously sent to a VLM for multimodal reasoning and also stored in a structured \emph{Case Memory Bank}.
This abstraction enables consistent cross-phase understanding while preserving diagnostic uncertainty and clinical semantics.
\textbf{(2) Knowledge-Guided Diagnostic Control Module (KDCM).}
As shown in Fig.~\ref{fig:pdct_agent} (b), to model clinical phase-selection policies, we construct a \emph{Phase-Aware Diagnostic Knowledge Database}, extracted from multi-phase radiology reports to encode structured diagnostic policies regarding \textit{phase sufficiency.}
Within the agent loop, the VLM receives Evidence Packages, retrieves historical evidence from the Case Memory Bank, queries the knowledge database, and determines whether current information is diagnostically sufficient.
If not sufficient, PD-CTAgent requests an additional CT phase, which re-enters the CSAM for iterative reasoning.
The process continues until diagnostic sufficiency is reached and a final report is generated.
The policy-driven agent design further allows PD-CTAgent to flexibly follow different institutional, regional, or guideline-specific diagnostic protocols.
By modeling CT phase selection as an iterative, knowledge-guided decision process, PD-CTAgent bridges the gap between static CT analysis and real-world clinical workflows.
Experimental results on two public datasets and one private dataset (LIDC dataset\cite{armato2011lung}, MCT-LTDiag dataset \cite{wu2025multi}, and Kidney53) demonstrate the superior performance of PD-CTAgent. Code will be made publicly available upon acceptance.

\textbf{Our contributions are summarized as follows:}

\begin{itemize}
\item We propose PD-CTAgent, an agent-based framework that models CT phase selection and diagnostic reasoning as an iterative evidence sufficiency problem, bridging the gap between static CT analysis and real-world clinical workflows.

\item We design Clinical Structure Abstraction Module (CSAM), which constructs structured Evidence Packages from multi-phase CT scans, enabling consistent cross-phase representation while preserving uncertainty and anatomical structure.

\item We introduce Knowledge-Guided Diagnostic Control Module (KDCM), which integrates retrieval-augmented clinical knowledge with rule-constrained decision policies to support evidence-driven phase escalation and diagnostic reasoning.

\item Extensive experiments on three datasets (MCT-LTDiag, Kidney53, and LIDC) demonstrate that PD-CTAgent consistently outperforms existing general VLMs, medical VLMs, and agentic baselines on both diagnostic accuracy and phase conversion tasks.
\end{itemize}

\section{Method}
\begin{figure*}[t]
    \centering
    \includegraphics[width=\textwidth]{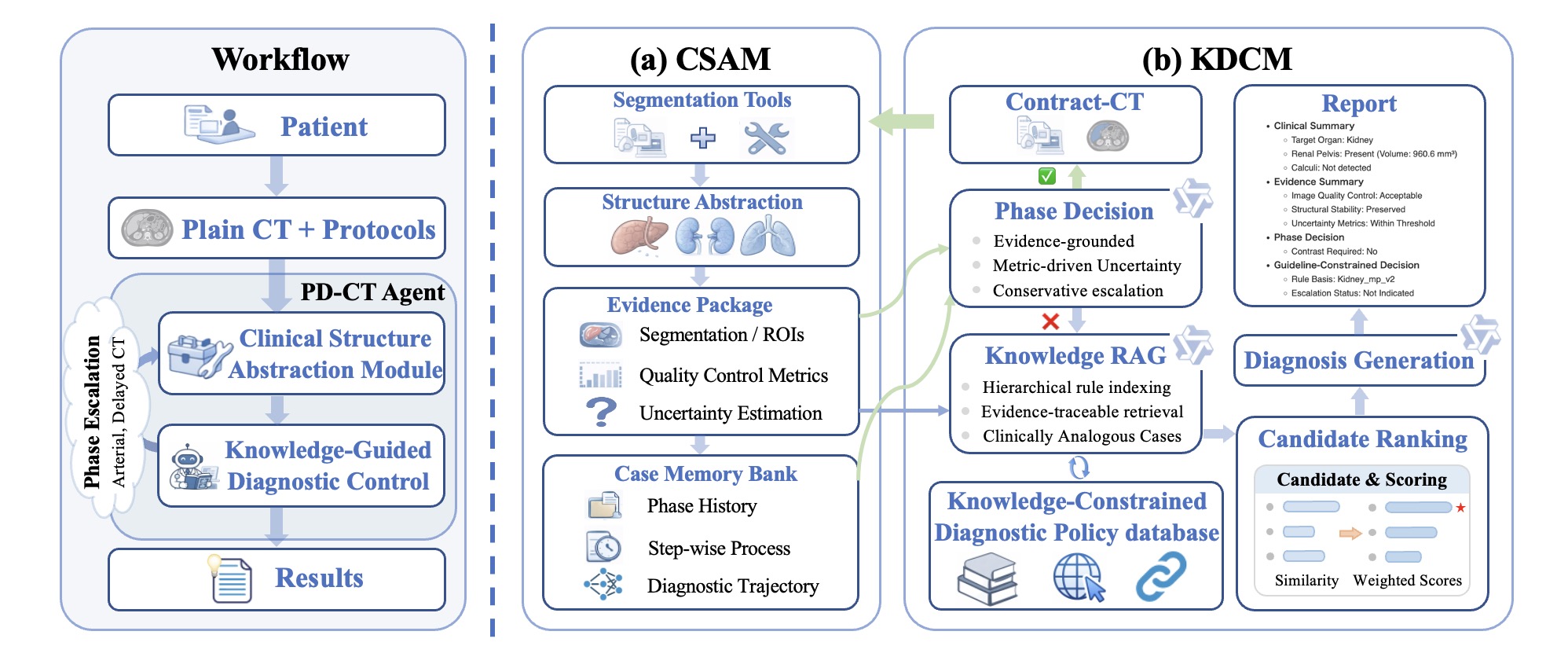}
    \vspace{-0.4cm}
    \caption{The overall architecture of PD-CTAgent. The framework consists of a Clinical Structure Abstraction Module (CSAM) and a Knowledge-Guided Decision-Making Module (KDCM), enabling iterative evidence extraction, phase escalation, and diagnostic reasoning from CT scans.}
    \label{fig:pdct_agent}
\end{figure*}

% \subsection{Overview}
As in Fig.~\ref{fig:pdct_agent}, we propose \textbf{PD-CTAgent}, an agent-based framework integrating structure-aware perception (CSAM) and knowledge-guided reasoning (KDCM). 
The agent first process an initial plain CT scan $I_0$. At each iteration $t$, CSAM extracts an evidence package $E_t = (F_t, U_t)$ from the current CT scan $I_t$, and KDCM evaluates whether diagnostic sufficiency is achieved. 
If insufficiency is detected, a new CT phase is acquired ($I_{t+1}$) and the loop continues. The final diagnostic report $R$ is generated only once sufficiency is reached:

\begin{equation}
\begin{aligned}
I_0 &\xrightarrow{\text{CSAM}} E_0 \xrightarrow{\text{KDCM}} I_1 \\
I_t &\xrightarrow{\text{CSAM}} E_t \xrightarrow{\text{KDCM}}
\begin{cases}
I_{t+1}, & \text{if insufficiency} \\
R, & \text{if sufficient}
\end{cases}
\end{aligned}
\end{equation}

Within CSAM, we introduce an organ-aware tool routing mechanism to support multi-organ scenarios, and a structure-aware uncertainty modeling framework to guide phase escalation. 
Within KDCM, we implement a guideline-driven decision-making framework, including offline guideline-to-DSL compilation, structured retrieval-augmented generation (RAG), and rule--LLM controlled fusion. 
Accordingly, we next introduce CSAM and KDCM in detail.

\subsection{Clinical Structure Abstraction Module (CSAM)}
\subsubsection{Organ-aware Tool Routing.}

Given the current CT scan $I_t$ (with $I_0$ being plain CT), we perform organ-level segmentation to obtain structured findings $F_t$. 
Rather than feeding raw images into the VLM, $I_t$ is converted into a structured representation, enabling organ-aware routing to disease-specific diagnostic tools and cross-organ reasoning. 

The resulting outputs form the evidence package 
$E_t=(F_t,U_t,Q_t)$, 
where $F_t$ denotes anatomical and lesion findings, $U_t$ uncertainty metrics, and $Q_t$ quality-control indicators. 
Specifically, \(F_t\) consists of structured anatomical and lesion-level descriptors, including the target organ, available CT phase, segmentation mask statistics, lesion location, lesion size or volume, organ-level abnormality flags, and disease-specific tool outputs. $U_t$ contains uncertainty-related measurements such as boundary uncertainty, critical-region uncertainty, and multiple-inference consistency. $Q_t$ records image quality and validity indicators. This structured representation allows the VLM to reason over compact clinical evidence rather than raw image slices.

The evidence package, together with a structured prompt and lightweight retrieval context, is fed into KGDC for phase-sufficiency evaluation. 
This abstraction suppresses irrelevant visual signals and reduces hallucination risk, enabling grounded and traceable phase-escalation decisions.
\subsubsection{Uncertainty-Guided Phase Escalation.}

Uncertainty-Guided Phase Escalation determines whether an additional CT phase is required based on structure-aware predictive instability. We design three complementary indicators.

\paragraph{Signed Distance Uncertainty (SDU).}
SDU measures boundary-focused predictive uncertainty. Let $M:\Omega \to \{0,1\}$ denote a segmentation mask, and define the foreground region as $\Omega_M = \{\mathbf{x} \mid M(\mathbf{x})=1\}$.

The signed distance function (in mm) is defined as:
\begin{equation}
d(\mathbf{x}) =
\begin{cases}
-\mathrm{dist}(\mathbf{x}, \partial \Omega_M), & \mathbf{x} \in \Omega_M \\
\ \mathrm{dist}(\mathbf{x}, \partial \Omega_M), & \mathbf{x} \notin \Omega_M
\end{cases}
\end{equation}

We define an adaptive boundary band:
\begin{equation}
B = \{\mathbf{x} \mid |d(\mathbf{x})| \le r_b\}.
\end{equation}

Given voxel-wise entropy
\begin{equation}
H(\mathbf{x}) = -\sum_c p_c(\mathbf{x}) \log p_c(\mathbf{x}),
\end{equation}

SDU is computed as:
\begin{equation}
\mathrm{SDU} = \frac{1}{|B|} \sum_{\mathbf{x} \in B} H(\mathbf{x}).
\end{equation}

\paragraph{Critical-Region Uncertainty.}
To focus on diagnostically influential regions, we define a decision-critical shell $\Omega_c=\{\mathbf{x}\mid |d(\mathbf{x})|\le r_c\}$ around the structure. 
Uncertainty is computed as the average entropy within $\Omega_c$, emphasizing regions most relevant to clinical decisions.

\paragraph{Multiple Inference Consistency (MIC).}
MIC evaluates the stability of stochastic segmentation predictions. Given $k$ repeated inferences $\{M_i\}_{i=1}^k$, MIC is defined as the average pairwise Dice similarity:

\begin{equation}
\mathrm{MIC} =
\frac{2}{k(k-1)} \sum_{1 \le i < j \le k}
\mathrm{Dice}(M_i, M_j).
\end{equation}

In practice, the repeated inferences are generated using lightweight test-time perturbations, including random seed variation, intensity jittering, and small spatial augmentations. MIC therefore measures whether the segmentation output remains stable under minor input and inference perturbations. A low MIC value indicates unstable structural evidence and can trigger conservative phase escalation.

The quality-control term $Q_t$ includes slice completeness, anatomical coverage, motion artifact flag, and signal-to-noise quality. A scan is marked as low-quality when the target anatomy is incompletely covered, severe motion artifacts are detected, or slice continuity is insufficient. These quality-control flags are included in the Evidence Package and can trigger conservative phase escalation when the current phase is unreliable.

\subsubsection{Case Memory Bank.}\label{sec:case memory bank}

To support iterative phase escalation, we maintain a structured Case Memory Bank that records the diagnostic trajectory across iterations. 
At iteration $t$, the system state is summarized as $S_t=(I_t,E_t,D_t,\mathcal{G}_t)$, where $I_t$ is the active CT phase, $E_t$ the evidence package, $D_t$ the policy decision, and $\mathcal{G}_t$ the matched guideline trace. 

The case memory stores the compact trajectory $\mathcal{M}_{\text{case}}=\{S_0,\dots,S_T\}$ without raw volumetric data for efficiency. 
A complementary skill memory $\mathcal{M}_{\text{skill}}=\{\theta\}$ maintains reusable calibration priors (e.g., uncertainty thresholds). 
At each step, KGDC operates on $S_t$ while optionally referencing $\mathcal{M}_{\text{case}}$ for consistency-aware decisions.

\subsection{Knowledge-Guided Diagnostic Control Module}

\subsubsection{Knowledge-Constrained Policy Database.}

Clinical guideline text \(G\) is compiled offline into structured assets \(K=\{R,E,O\}\), where \(R\) denotes rule-level chunks, \(E\) evidence fragments, and \(O\) a terminology ontology. Each rule-level chunk stores the original clinical text, structured metadata (e.g., organ, disease, imaging finding, phase condition, and evidence requirement), and an embedding vector for semantic retrieval. Each rule is linked to valid evidence entries, ensuring structural consistency. The compiled knowledge is stored locally for online retrieval.

The offline compilation procedure contains three steps. First, clinical guideline texts, CT protocol descriptions, and representative radiology reports are filtered by organ, disease, phase, and evidence-related keywords. Second, relevant paragraphs are converted into rule-level entries, where each entry contains the target organ, imaging finding, phase condition, evidence requirement, recommended action, and source identifier. Third, the extracted rules are manually checked, linked to evidence fragments and ontology terms, and encoded into embeddings for retrieval. This procedure avoids relying on free-form long-context reasoning during inference and makes each phase-escalation decision traceable to structured knowledge assets.

% Clinical guideline text $G$ is compiled offline into structured assets $K=\{R,E,O\}$, where $R$ denotes executable rules, $E$ evidence fragments, and $O$ a terminology ontology. 
% Each rule is linked to valid evidence entries, ensuring structural consistency. 
% The compiled knowledge is stored locally for online retrieval.

% The offline compilation procedure contains three steps. First, clinical guideline texts, CT protocol descriptions, and representative radiology reports are filtered by organ, disease, phase, and evidence-related keywords. Second, relevant paragraphs are converted into rule-level entries, where each entry contains the target organ, imaging finding, phase condition, evidence requirement, recommended action, and source identifier. Third, the extracted rules are manually checked and linked to evidence fragments and ontology terms for retrieval. This procedure avoids relying on free-form long-context reasoning during inference and makes each phase-escalation decision traceable to structured knowledge assets.

\subsubsection{Structured Retrieval-Augmented Reasoning.}

Given the current state \(S_t\), the Evidence Package is converted into a structured query containing organ, phase, lesion type, and evidence fields. We first perform organ- and phase-aware rule filtering to obtain \(R_t \subset R\), and then retrieve supporting evidence \(E_t^{ret}=\mathrm{TopK}(q(S_t),E)\), where \(q(\cdot)\) is a semantic query constructed from lesion morphology, density abnormality, boundary ambiguity, image-quality flags, and uncertainty findings. Ontology \(O\) is used for label normalization. All knowledge assets are locally loaded, avoiding reliance on external services.

% Given the current state $S_t$, we perform organ-aware rule filtering to obtain $R_t\subset R$, and retrieve supporting evidence $E_t^{\text{ret}}=\mathrm{TopK}(q(S_t),E)$, where $q(\cdot)$ is a structured query function. 
% Ontology $O$ is used for label normalization. 
% All knowledge assets are locally loaded, avoiding reliance on external services.

\subsection{VLM-Guided Strategy Generation}

At iteration $t$, the policy module determines whether phase escalation is required based on the current state $S_t$.

A rule-based decision $d_r \in \{0,1\}$ is first derived from predefined clinical criteria $R_t$, while a VLM-based decision $d_v \in \{0,1\}$ is generated from the joint input of the current state $S_t$ and retrieved evidence $E_t^{\text{ret}}$.

The final decision is obtained via a rule-prioritized fusion strategy:
\begin{equation}
D_t =
\begin{cases}
1, & \text{if } d_r = 1, \\
d_v, & \text{otherwise}.
\end{cases}
\end{equation}

This design ensures conservative escalation behavior, allowing escalation when either rule-based or model-based signals indicate necessity, while preventing unsupported de-escalation from $1 \rightarrow 0$.

\begin{table}[t]
\centering
\caption{Dataset and phase/evidence characterization.}
\label{tab:dataset_phase}
\renewcommand{\arraystretch}{1.2}
\resizebox{\columnwidth}{!}{
\begin{tabular}{lccc}
\toprule
Dataset & Organ & Cases & Phase / Evidence Sources \\
\midrule
LIDC-IDRI & Lung & 2630 nodules & Plain CT / reader masks \\
MCT-LTDiag & Liver & 60 & NC, arterial, PVP, delayed \\
Kidney53 & Kidney & 60 & Plain and contrast/excretory CT \\
Kidney paired subset & Kidney & 46 & Paired plain/excretory evidence packages \\
\bottomrule
\end{tabular}
}
\end{table}

\begin{table*}[!t]
\centering
\setlength{\tabcolsep}{1.8mm}
\caption{Comparison of general VLMs, medical VLMs, and medical agentic systems on MCT, Kidney, and LIDC datasets.}
\label{tab:result}

\begin{tabular}{l|cc|cc|cc|cc|cc}
\toprule

\multirow{2}{*}{Methods}
& \multicolumn{2}{c|}{MCT-LTDiag (Diag)}
& \multicolumn{2}{c|}{MCT-LTDiag (Conv)}
& \multicolumn{2}{c|}{Kidney53 (Diag)}
& \multicolumn{2}{c|}{Kidney53 (Conv)}
& \multicolumn{2}{c}{LIDC (Diag)} \\

& Acc & F1 & Acc & F1 & Acc & F1 & Acc & F1 & Acc & F1 \\

\midrule

GPT-5.2-Pro\cite{openai_gpt5.2}
& 51.7 & 0.34 & 65.0 & 0.65
& 72.6 & 0.68 & 20.8 & 0.29
& 61.8 & 0.71 \\

InternVL3.5-8B\cite{internvl3_5}
& 56.7 & 0.36 & 15.9 & 0.26
& 51.9 & 0.51 & 0.0 & 0.00
& 28.6 & 0.19 \\

Qwen2.5-VL-7B\cite{qwen2_5vl}
& 25.0 & 0.20 & 6.7 & 0.13
& 46.2 & 0.45 & 5.7 & 0.09
& 34.2 & 0.33 \\

\midrule

Lingshu\cite{xu2025lingshu}
& 72.4 & 0.42 & 9.1 & 0.09
& 53.9 & 0.54 & 5.7 & 0.07
& 39.0 & 0.44 \\

MedVLM-R1\cite{pan2025medvlm}
& 21.7 & 0.18 & 0.0 & 0.00
& 48.1 & 0.42 & 0.0 & 0.00
& 28.6 & 0.20 \\

\midrule

MedAgent-Pro\cite{wang2025medagent}
& 36.7 & 0.32 & -- & --
& 36.0 & 0.52 & -- & --
& 73.5 & 0.74 \\

PD-CT Agent (Ours)
& \textbf{76.7} & \textbf{0.69} & \textbf{83.3} & \textbf{0.79}
& \textbf{65.4} & 0.40 & \textbf{56.6} & \textbf{0.69}
& \textbf{77.0} & \textbf{0.87} \\

\bottomrule
\end{tabular}
\end{table*}

\subsection{Protocols-Aware Agent Reasoning}

PD-CT Agent accepts an active hospital rule set as part of the agent context. The rule set specifies the diagnostic scenario, required evidence fields, phase-selection criteria, report emphasis, and next-step recommendations. During inference, the agent conditions the entire workflow on this rule set rather than only using it as a post-hoc decision constraint. Given the current Evidence Package and case memory, the agent retrieves the rules applicable to the target organ, disease scenario, and available CT phase. It then checks whether the required evidence fields are present, determines whether additional phase evidence is needed, and generates a rule-conditioned diagnostic report. Thus, the same case can keep the same phase decision when evidence
insufficiency is shared, while producing different diagnostic report emphases under different hospital rules.

\section{Experiments}

\subsection{Experimental Setup}

We evaluate PD-CTAgent on three datasets: \textbf{LIDC-IDRI}~\cite{armato2011lung} (plain lung CT, $512\times512$), \textbf{MCT-LTDiag}~\cite{wu2025multi} (multi-phase liver CT, $512\times512$) and \textbf{Kidney53} (private multi-phase kidney CT, 60 cases, $512\times512$).

Table~\ref{tab:dataset_phase} summarizes the phase characteristics and evidence-source settings of the evaluated datasets. MCT-LTDiag provides multi-phase liver CT scans, including non-contrast, arterial, portal venous, and delayed phases. Kidney53 consists of private multi-phase kidney CT cases, with the controlled workflow analysis further using 46 paired plain/excretory cases and cached 3D evidence packages. LIDC-IDRI is a plain CT dataset, serving as a single-phase diagnostic setting with reader-derived evidence.

Diagnostic and phase-escalation performance is measured by Acc and F1. Baselines include general VLMs (GPT-5.2-Pro~\cite{openai_gpt5.2}, Qwen2.5-VL-7B~\cite{qwen2_5vl}, InternVL3.5-8B~\cite{internvl3_5}), medical VLMs (Lingshu~\cite{xu2025lingshu}, MedVLM-R1~\cite{pan2025medvlm}), and a medical agent (MedAgent-Pro~\cite{wang2025medagent}; diagnostic only). PD-CT Agent is implemented on LangGraph with each module as a node, uses Qwen2.5-7B-VL as VLM, and loads offline-compiled knowledge assets (rules, evidence, ontology) locally for interpretable, controllable inference.

\subsection{Quantitative Evaluation}
As in Table~\ref{tab:result},  We evaluate PD-CTAgent on two tasks: (1) diagnostic prediction and (2) phase-escalation decision, which measures whether the model correctly determines the need for additional CT phases. 

On MCT-LTDiag, general VLMs achieve moderate diagnostic accuracy (GPT-5.2-pro 51.7\%) but perform poorly on phase conversion (Acc $< 66\%$), whereas PD-CTAgent reaches 76.7\% diagnostic and 83.3\% conversion accuracy, demonstrating effective lesion detection and contrast-phase triggering.

% On Kidney53, some baselines (InternVL2.5-8B, Qwen-2.5-VL-7B, MedVLM) struggle with limited contrast information, and PD-CTAgent improves both metrics. However, its diagnostic F1 is lower than MedAgent-Pro (0.68) and slightly below Lingshu (0.54), reflecting challenges in lesion boundary delineation under small-sample multi-phase conditions. On LIDC (plain CT only), PD-CTAgent leads in both diagnostic Acc (77.0\%) and F1 (0.87), confirming reliable single-phase diagnosis. 

On Kidney53, several baselines, including InternVL2.5-8B, Qwen2.5-VL-7B, and MedVLM, show limited performance under incomplete contrast information. PD-CTAgent improves both diagnostic accuracy and phase-escalation performance by incorporating structured evidence and knowledge-guided phase control. Although its diagnostic F1 remains lower than MedAgent-Pro (0.68) and slightly below Lingshu (0.54), this
result reflects the difficulty of lesion boundary characterization in a small-sample multi-phase setting. On LIDC, which contains plain CT only, PD-CTAgent achieves the best diagnostic accuracy (77.0\%) and F1 score (0.87), indicating that the proposed framework can also perform reliable diagnosis when no phase escalation is required.

Overall, PD-CTAgent consistently balances accurate diagnosis and robust phase-escalation, while remaining sensitive to dataset-specific challenges where sample size or contrast quality limits performance.

\subsection{Qualitative Evaluation and Visualization}

% Fig.~\ref{fig:case} shows two representative Kidney53 cases illustrating PD-CTAgent's phase-selection behavior. In the left case, the renal anatomy is clearly visible on the plain CT, and no calculi are present, so no phase escalation occurs. In the right case, the lesion is unclear on plain CT but highlighted in contrast-enhanced images; PD-CTAgent correctly triggers phase escalation, enabling accurate lesion localization and final diagnosis. These examples demonstrate that phase-selection is evidence-driven and improves diagnostic reliability.

\begin{figure*}[!t]
    \centering
    \vspace{-0.5cm}
    \includegraphics[width=0.87\textwidth]{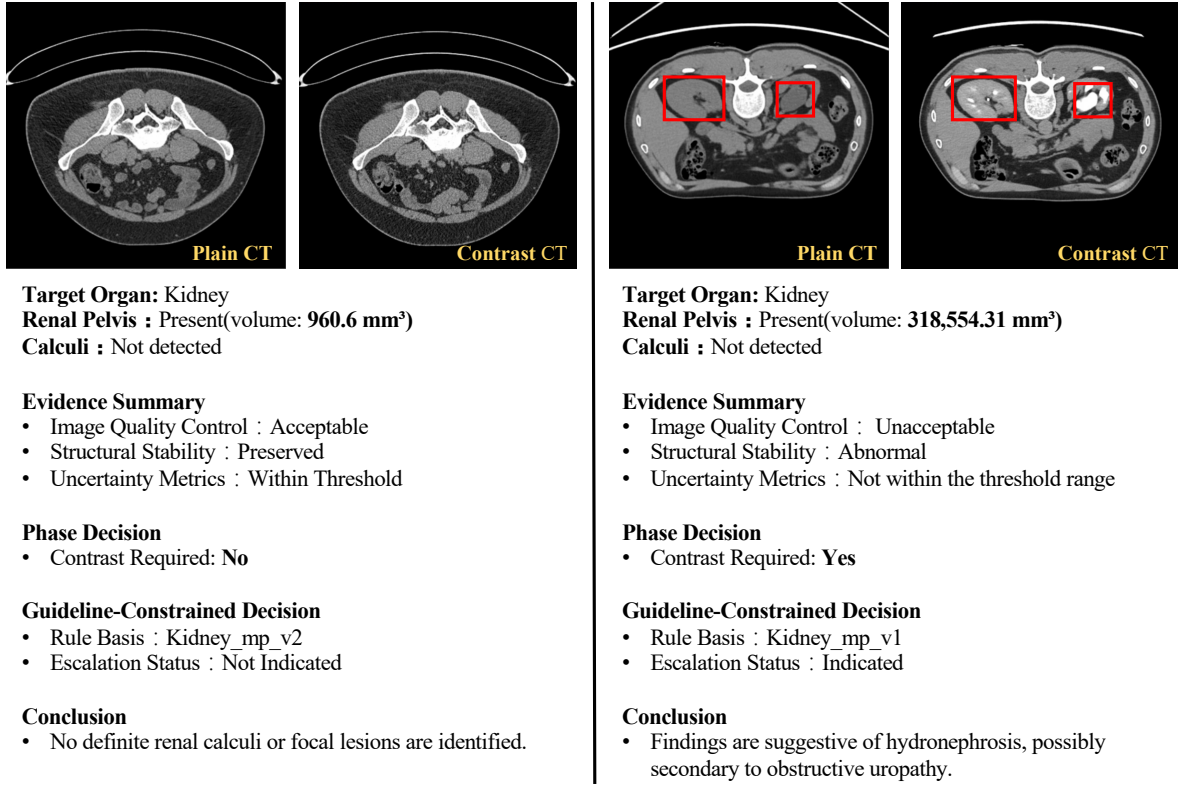}
    \caption{Qualitative examples of PD-CTAgent on Kidney53. The left case shows sufficient plain-CT evidence without phase escalation, whereas the right case illustrates evidence-driven escalation to contrast-enhanced CT for clearer lesion localization.}
    \label{fig:case}
\end{figure*}

Fig.~\ref{fig:case} shows two representative Kidney53 cases illustrating PD-CTAgent's phase-selection behavior. In the left case, the renal anatomy is clearly visible on the plain CT, and no calculi are detected. The extracted evidence package provides sufficient anatomical and lesion-related information, so the agent determines that the current phase is adequate and no phase escalation is required. In the right case, the lesion is unclear on plain CT and the current evidence is insufficient for reliable diagnosis. PD-CTAgent therefore requests an additional contrast-enhanced phase, where the lesion becomes more distinguishable and can be localized more accurately. These examples show that the phase-selection process is driven by the sufficiency and reliability of structured evidence rather than by a fixed escalation rule, supporting improved diagnostic reliability while avoiding unnecessary additional phases.

\begin{table*}[!t]
\centering
\renewcommand{\arraystretch}{1.1}
\setlength{\tabcolsep}{1.8mm}
\caption{Ablation studies on key components of PD-CTAgent on MCT-LTDiag and Kidney53 datasets.}
\label{tab:ablation}

\begin{tabular}{ccc|cc|cc|cc|cc}
\toprule

Evidence & Uncertainty & Knowledge
& \multicolumn{4}{c|}{MCT-LTDiag}
& \multicolumn{4}{c}{Kidney53} \\

& & 
& \multicolumn{2}{c}{Diag}
& \multicolumn{2}{c|}{Conv}
& \multicolumn{2}{c}{Diag}
& \multicolumn{2}{c}{Conv} \\

& & 
& Acc & F1 & Acc & F1 & Acc & F1 & Acc & F1 \\

\midrule

& $\checkmark$ & $\checkmark$
& 33.3 & 0.26 & 55.0 & 0.18 & 34.6 & 0.26 & 49.1 & 0.63 \\

$\checkmark$ & & $\checkmark$
& \underline{38.3} & \underline{0.30} & 45.2 & 0.27 & \underline{55.8} & \underline{0.47} & \textbf{60.4} & \underline{0.64} \\

$\checkmark$ & $\checkmark$ &
& 37.4 & 0.28 & \underline{58.3} & \underline{0.39} & 51.0 & \textbf{0.49} & 37.7 & 0.44 \\

$\checkmark$ & $\checkmark$ & $\checkmark$
& \textbf{76.7} & \textbf{0.69} & \textbf{83.3} & \textbf{0.79} & \textbf{65.4} & 0.40 & \underline{56.6} & \textbf{0.69} \\

\bottomrule
\end{tabular}
\vspace{-0.2cm}
\end{table*}

\subsection{Ablation Studies}

Table~\ref{tab:ablation} shows that each component contributes to performance. On MCT-LTDiag, adding the evidence package improves diagnostics (33.3\% → 38.3\% Acc), uncertainty boosts conversion (55.0\% → 58.3\% Acc), and the knowledge base aligns diagnostics with phase selection; the full model reaches 76.7\%/83.3\% (diagnostic/conversion Acc). On Kidney53, evidence and uncertainty improve metrics, but only the full model balances diagnostic and conversion accuracy (65.4\%/56.6\%), highlighting their complementary roles.

\begin{table*}[t]
\centering
\caption{Controlled analysis of PD-CTAgent workflow. Avg. Evidence denotes the average number of CT phases or evidence sources used. Task Score is pelvis Dice for kidney, CNR for liver, and consensus Dice for lung. Missed Needed denotes the proportion of cases where additional evidence was needed but not requested.}
\label{tab:controlled_analysis}
\renewcommand{\arraystretch}{1.2}
\resizebox{\textwidth}{!}{
\begin{tabular}{ll l c c c c l}
\hline
Analysis & Dataset & Method & Avg. Evidence & Task Score & Phase/Esc. F1 & Missed Needed & Interpretation \\
\hline
Initial evidence sufficiency & Kidney paired CT & Plain one-shot
& 1.000 & Dice 0.3715 & 0.0000 & 1.0000
& Initial phase alone is insufficient for many cases. \\
Initial evidence sufficiency & Kidney paired CT & Full adaptive policy
& 1.978 & Dice 0.7434 & 0.9048 & 0.0256
& Additional phase evidence improves structured diagnosis. \\
Direct VLM planning & Kidney paired CT & Qwen cached planner
& 1.739 & Dice 0.5981 & 0.8219 & 0.2308
& Direct VLM planning misses many needed escalations. \\
Iterative re-entry & Kidney paired CT & w/o iterative re-entry
& 1.978 & Dice 0.3715 & 0.9048 & 0.0256
& Same escalation decisions fail without re-processing new evidence. \\
Iterative re-entry & Kidney paired CT & Full adaptive policy
& 1.978 & Dice 0.7434 & 0.9048 & 0.0256
& Re-entering CSAM after phase acquisition is necessary. \\
Strong static baseline & Kidney paired CT & Fixed all evidence
& 2.000 & Dice 0.7465 & 0.9176 & 0.0000
& Fixed-all is strong but does not model phase sufficiency. \\
Multi-organ instantiation & Liver CT & Oracle single phase
& 1.817 & CNR 2.3441 & 0.9684 & 0.0000
& Liver phase selection has strong potential. \\
Multi-organ instantiation & Liver CT & Adaptive CV threshold
& 3.950 & CNR 2.3236 & 0.8571 & 0.0217
& Current deployable liver policy remains conservative. \\
Multi-organ instantiation & Lung CT & ExtraTrees adaptive
& 3.706 & Dice 0.9996 & 0.5447 & 0.0014
& Workflow can be instantiated for reader-consensus evidence. \\
Multi-organ instantiation & Lung CT & Fixed all evidence
& 4.000 & Dice 1.0000 & 0.8553 & 0.0000
& Full evidence upper baseline. \\
\hline
\end{tabular}
}
\end{table*}

\subsection{Controlled Analysis of PD-CTAgent Workflow}

In addition to the main diagnostic and phase-conversion evaluation, we conduct controlled analyses to further clarify the design choices of PD-CTAgent. These analyses are not intended to redefine the task, but to examine whether the key elements of the proposed workflow are functionally necessary. Specifically, we evaluate four questions: whether the initial phase is sufficient, whether direct VLM planning can replace the agent workflow, whether newly acquired phases must re-enter CSAM, and whether the same workflow can be instantiated beyond urinary CT.

In this analysis, \emph{iterative re-entry} refers to the process in which a newly acquired CT phase is not simply appended to the language context. Instead, it is sent back to the Clinical Structure Abstraction Module (CSAM), where segmentation-derived features, quality-control signals, uncertainty indicators, and the Evidence Package are regenerated. This mechanism ensures that phase-escalation decisions update the structured imaging evidence used for final diagnosis.

\begin{figure*}[!t]
    \centering
    \includegraphics[width=0.97\textwidth]{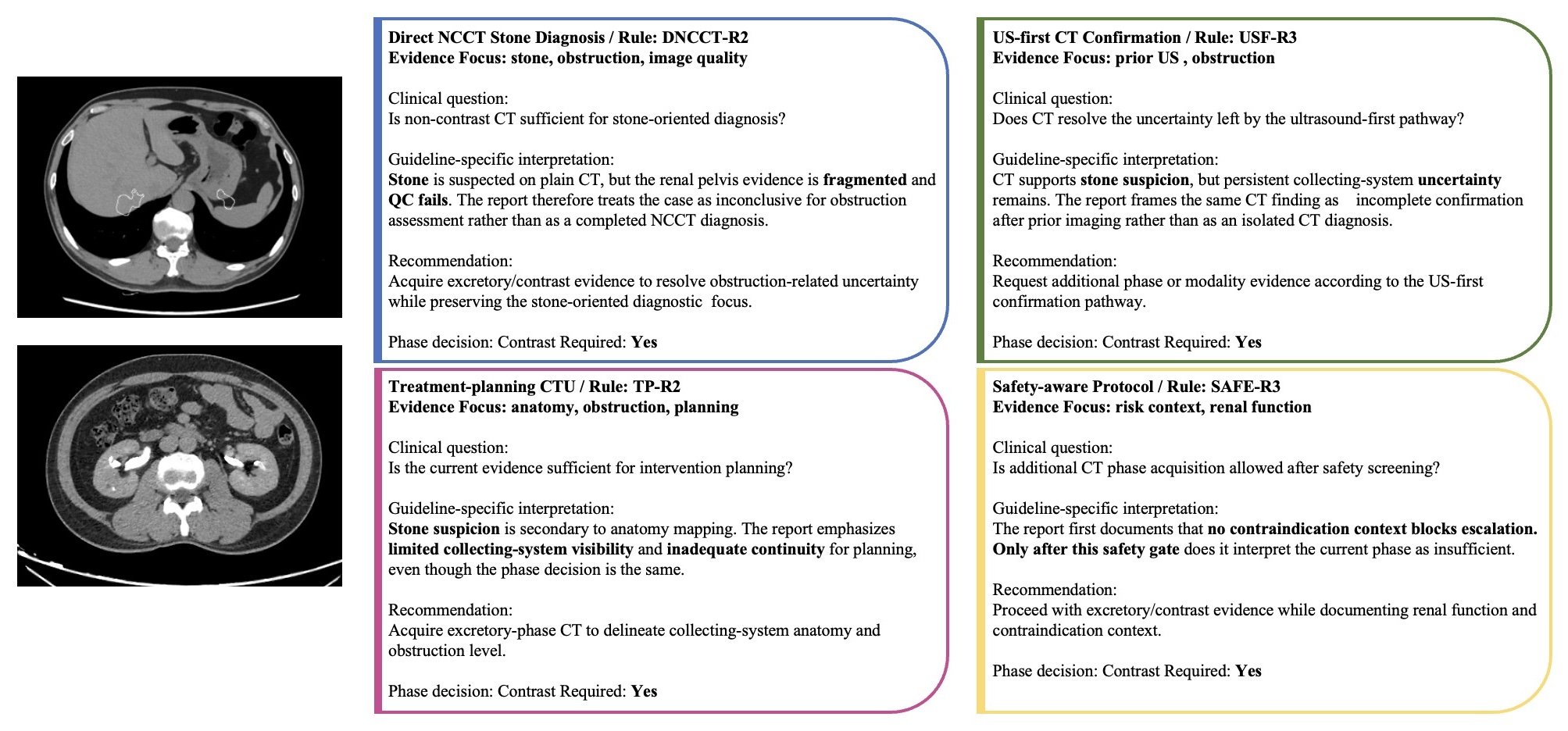}
    \vspace{-0.1cm}
    \caption{Case-level example of guideline-conditioned reasoning with four hospital-specific rules. The same CT case is evaluated under DNCCT-R2 for direct non-contrast stone assessment, USF-R3 for ultrasound-first CT confirmation, TP-R2 for treatment-planning CTU, and SAFE-R3 for safety-aware CT escalation. All four rules lead to the same phase decision because the plain CT evidence is structurally insufficient. However, the generated reports differ in diagnostic emphasis: DNCCT-R2 focuses on incomplete stone-oriented assessment, USF-R3 frames the CT as confirmatory evidence after prior imaging, TP-R2 emphasizes collecting-system anatomy for planning, and SAFE-R3 documents safety-context verification before escalation.}
    \label{fig:rule}
\end{figure*}

Table~\ref{tab:controlled_analysis} summarizes the controlled analysis results. On paired kidney CT, plain one-shot inference obtains a low pelvis Dice of 0.3715 and a missed-needed rate of 1.0000, indicating that the initial plain phase is insufficient for many cases. The full adaptive workflow improves Dice to 0.7434 and reduces the missed-needed rate to 0.0256. The Qwen cached planner reduces average phase use to 1.739 but has a missed-needed rate of 0.2308, suggesting that direct VLM planning is unstable for phase-escalation decisions without structured control.

Most importantly, removing iterative re-entry keeps the same average evidence and escalation F1 as the full adaptive policy, but reduces pelvis Dice from 0.7434 to 0.3715. This confirms that newly acquired phases must be re-processed by CSAM rather than only considered at the language-reasoning level. The liver and lung rows further show that the same workflow can be instantiated beyond urinary CT. For liver CT, oracle single-phase selection reaches the same CNR as all-phase evidence using fewer phases, suggesting that phase selection has strong potential, although the current deployable liver policy remains conservative. For lung CT, ExtraTrees adaptive control reaches Dice 0.9996 using 3.706 evidence sources, close to the fixed-all Dice of 1.0000. These results indicate that the PD-CTAgent workflow can be instantiated beyond urinary CT, while fixed-all evidence remains a strong upper baseline in some settings.

Overall, the controlled analysis supports the necessity of the PD-CTAgent workflow without changing the original diagnostic task. The results do not imply that an LLM-based agent universally outperforms all fixed-evidence baselines. Instead, they show that structured evidence extraction, knowledge-guided control, and iterative perception re-entry address complementary limitations of direct VLM reasoning and static single-phase inference.

\subsection{Guideline-Conditioned Case Study}

% For guideline-conditioned reasoning, each hospital protocol is compiled into rule-level chunks. Each rule contains a diagnostic scope, required evidence fields, a matching condition, and a recommended next action. During inference, the current Evidence Package is converted. As in Fig.~\ref{fig:rule}

For guideline-conditioned reasoning, each hospital protocol is compiled into rule-level chunks with diagnostic scope, evidence
requirements, matching conditions, and recommended actions.
During inference, the current Evidence Package is matched with the active hospital-specific rules. As shown in Fig.~\ref{fig:rule}, the same CT case can follow different guideline-driven reasoning pathways, such as non-contrast assessment, collecting-system confirmation, or treatment-planning evaluation. Although the intermediate analysis and report emphasis vary across protocols, the final phase decision and diagnostic conclusion remain consistent because they are grounded in the same image-derived evidence.

\section{Conclusion}

In this work, we presented PD-CTAgent, a policy-driven agent framework that models clinically consistent CT phase selection and diagnostic reasoning. 
Unlike conventional static CT AI systems, PD-CTAgent formulates phase acquisition as an iterative, knowledge-guided decision process. 
By integrating structured visual abstraction with knowledge-based diagnostic control, the proposed framework enables uncertainty-aware escalation and workflow-aligned reasoning across heterogeneous CT phases. 
Experimental results on two public benchmarks and one private dataset demonstrate that PD-CTAgent effectively improves diagnostic reliability while reducing unnecessary multi-phase acquisition. 
These findings suggest that policy-driven agent modeling provides a promising direction toward workflow-aware and clinically grounded medical imaging AI systems.

\vspace{12pt}
% \color{red}
% IEEE conference templates contain guidance text for composing and formatting conference papers. Please ensure that all template text is removed from your conference paper prior to submission to the conference. Failure to remove the template text from your paper may result in your paper not being published.

\bibliographystyle{IEEEtran}
\bibliography{Reference}

\end{document}